\documentclass[10pt,twocolumn,letterpaper]{article}

\usepackage{cvpr}              
\usepackage{tabularx}
\usepackage[table]{xcolor} 
\usepackage{multirow}
\usepackage{pifont}
\usepackage{graphicx}
\usepackage{float}
\newcommand{\cmark}{\ding{51}} 
\newcommand{\xmark}{\ding{55}} 

\usepackage{listings}
\lstdefinestyle{mystyle}{
    backgroundcolor=\color{lightgray},
    keywordstyle=\color{black},
    numberstyle=\tiny\color{black},
    stringstyle=\color{black},
    commentstyle=\color{black}, 
    keywordstyle=\color{black}, 
    stringstyle=\color{black}, 
    basicstyle=\ttfamily\footnotesize, 
    basicstyle=\ttfamily\footnotesize,
    breakatwhitespace=true,
    breaklines=true,
    postbreak=\mbox{\hspace{0pt}},
    captionpos=b,
    keepspaces=true,
    numbers=none,
    numbersep=5pt,
    showspaces=false,
    showstringspaces=false,
    showtabs=false,
    tabsize=2
}
\definecolor{cvprblue}{rgb}{0.21,0.49,0.74}
\usepackage[pagebackref,breaklinks,colorlinks,allcolors=cvprblue]{hyperref}

\def\paperID{37785} 
\def\confName{CVPR}
\def\confYear{2026}

\title{Memory-Centric Embodied Question Answering}

\author{%
    Mingliang Zhai\textsuperscript{\rm 1,2},
    Zhi Gao\textsuperscript{\rm 1},
    Yuwei Wu\textsuperscript{\rm 2,1},
    Yunde Jia\textsuperscript{\rm 2,1}, \\
    \small \textsuperscript{1}Beijing Key Laboratory of Intelligent Information Technology,\\
    \small School of Computer Science and Technology, Beijing Institute of Technology\\
    \small \textsuperscript{2}Guangdong Laboratory of Machine Perception and Intelligent Computing, Shenzhen MSU-BIT University \\
  \small \texttt{\{zhaimingliang,wuyuwei,jiayunde,zhi.gao\}@bit.edu.cn} \\
  \small \hyperlink{https://memory-eqa.github.io}{\texttt{memory-eqa.github.io}}
}

\begin{document}
\maketitle
\begin{abstract}

Embodied Question Answering (EQA) requires agents to autonomously explore and comprehend the environment to answer context-dependent questions.
Typically, an EQA framework consists of four components: a planner, a memory module, a stopping module, and an answering module. 
However, the memory module is utilized inefficiently in existing methods, as the information it stores is leveraged solely for the answering module.
Such a design may result in redundant or inadequate exploration, leading to a suboptimal success rate.
To solve this problem, we propose \textbf{MemoryEQA}, an EQA framework centered on memory, which establishes mechanisms for memory storage, update, and retrieval, allowing memory information to contribute throughout the entire exploration process.
Specifically, we convert the observation into structured textual representations, which are stored in a vector library following a fixed structure. 
At each exploration step, we utilize a viewpoint comparison strategy to determine whether the memory requires updating. 
Before executing each module, we employ an entropy-based adaptive retrieval strategy to obtain the minimal yet sufficient memory information that satisfies the requirements of different modules. 
The retrieved module-specific information is then integrated with the current observation as input to the corresponding module.
To evaluate EQA models' memory capabilities, we constructed the benchmark based on HM3D called \textbf{MT-HM3D}, comprising 1,587 question-answer pairs involving multiple targets across various regions, which requires agents to maintain memory of exploration-acquired target information.
Experimental results on HM-EQA, MT-HM3D, and OpenEQA demonstrate the effectiveness of our framework, where a 9.9\% performance gain on MT-HM3D compared to baseline models further underscores the memory capability’s pivotal role in solving complex tasks.
\end{abstract}
\section{Introduction}
\label{sec:intro}

Embodied Question Answering (EQA) \cite{das2018embodied,gordon2018iqa,wijmans2019embodied,jiang2025beyond} is a challenging task in robotics that requires agents to actively interact with dynamic and evolving environments to answer natural language questions.
Existing EQA approaches \cite{ren2024explore,yu2019multi,das2018neural,chen2023not,dai2024think,jiang2025beyond} are typically designed around a planner, which coordinates other modules including the memory, answering, and stopping modules. However, in these planner-centric methods, the memory module is often underutilized, whose information is only leveraged by the answering module, leading to inefficient exploration and inaccurate responses, thereby reducing overall task success rates.
For example, given complex tasks that require exploring multiple regions to gather information for reasoning, an agent must effectively retain and share past observations across all modules to enable efficient planning and exploration. Yet, planner-centric designs typically rely only on the agent’s current observation for planning and stopping decisions, resulting in redundant or insufficient exploration.
As illustrated in Figure~\ref{fig:case}, when answering a question “Are the sofas in the living room and the media room the same color?”, the agent must explore two objects (“sofa in the living room,” “sofa in the media room”) to perform comparative reasoning. In the planner-centric framework, since the planner lacks direct access to the memory module, crucial information collected during exploration remains inaccessible. Consequently, even after discovering that “the sofa in the living room is white,” the planner may inefficiently re-explore the same sofa or fail to complete the task, as this critical knowledge is unavailable during subsequent decision-making.

In this paper, we propose MemoryEQA, a memory-centric EQA framework that maintains a memory library and infuses memory information into all modules to guide reasoning, enabling efficient exploration and accurate responses.
Specifically, MemoryEQA consists of four components: a planner, a memory module, a stopping module, and an answering module. For the planner, we augment the widely used frontier-based planner with memory guidance, leveraging memory information to steer planning. The memory module is equipped with mechanisms for storage, update, and retrieval, allowing memory to guide all processes in EQA. For the stopping and answering modules, we use different natural language queries to retrieve relevant information from the memory module and perform reasoning with a multi-modal large language model.

In doing so, we need to address a key challenge in the memory module design, that is, the instability of long-term memory updates and retrievals in dynamic, complex environments.
Irrelevant memories may be retrieved and adversely affect exploration, planning, reasoning, and answering. In other words, long-term memory in embodied environments is easily corrupted by noisy updates and irrelevant retrievals. To address this, we propose a viewpoint-contrastive memory update rule for updating natural language memories, and an entropy-based adaptive retrieval strategy to select the most relevant memory information to the user query, mitigating the impact of redundant or irrelevant information on subsequent reasoning.

To evaluate the memory capability of EQA models, 
we built a benchmark based on HM3D~\cite{ramakrishnan2021habitat} with 1587 QA pairs, called MT-HM3D.
Specifically, we utilize the Habitat simulator \cite{szot2021habitat} to sample 5 to 10 images per scene with constrained distances between sampling points. 
We then prompt GPT-4o \cite{gpt4o} to generate questions about multiple targets across different regions, which require the agent to memorize all question-relevant information during exploration.
Finally, low-quality data filtering is performed through manual verification.
Our contributions are as follows:
\begin{itemize}
    \item We propose a memory-centric EQA framework that provides a multi-modal hierarchical memory with a viewpoint-contrastive memory update rule and an entropy-based adaptive retrieval strategy for efficient exploration and reasoning.
    \item We construct the benchmark MT-HM3D based on the HM3D, offering more complex QA pairs about multiple targets across different regions, effectively evaluating EQA models' memory capabilities.
    \item MemoryEQA achieves a 9.9\% improvement in success rate over baseline methods on MT-HM3D, demonstrating the importance of memory capability in solving complex tasks. Additionally, it sets a new state-of-the-art on the open-vocabulary EQA dataset OpenEQA and achieves comparable success rates to previous best methods on HM-EQA.
\end{itemize}

\section{Related Works}
\subsection{Embodied Question Answering}
Embodied question answering \cite{das2018embodied, gordon2018iqa, yu2019multi, cangea2019videonavqa,das2018neural} has become a challenging task for testing the robot's ability to autonomously plan tasks and establish semantic understanding of the environment.
Das \textit{et al.} \cite{das2018embodied} first proposed the EQA task and utilized a planner-controller navigation module to accomplish EQA task in virtual environments.
Allen \textit{et al.} \cite{ren2024explore} constructed a 2D semantic map of the environment for specific tasks to guide exploration. 
Answer \textit{et al.} \cite{anwar2024remembr} and Xie \textit{et al.} \cite{xie2024embodied} developed offline embodied modules that could be queried by large language models (LLMs). 
Arjun \textit{et al.} \cite{majumdar2024openeqa} utilized video memory to answer implicit questions using long-context Visal Language Models (VLMs).
Saumya \textit{et al.} \cite{saxena2024grapheqa} bridged the gap between semantic memory and planners by placing the planner within a compact scene representation. 
Different from existing methods, we position memory as the core module of the embodied question answering framework, enabling the planner, stopping criteria, and question-answering module to reason based on memory information. 
The memory-centric framework overcomes the limitations of planner-centric methods by infusing memory information into all modules, enabling more efficient exploration and accurate responses through enhanced inter-module interaction.

\subsection{Memory Mechanisms in Embodied Agent}
Early works utilized the hidden state of the LSTM \cite{graves2012long} navigation model as a dense memory representation of the scene \cite{szot2021habitat, wijmans2019dd}. However, the expressive power of this single state vector is limited, making it challenging to handle the storage of memory information in complex 3D real-world scenarios.
Recent approaches have modeled scene memory as 2D metric grid \cite{anderson2019chasing,blukis2018mapping,cartillier2021semantic,chaplot2020object}, topological map \cite{savinov2018semi,wu2019bayesian}, and 3D semantic map\cite{cheng2018geometry,prabhudesai2019embodied,tung2019learning,saxena2024grapheqa} to serve as global memory information. 
Although these methods have shown promising performance in navigation task, the complexity of indoor environments, coupled with the lack of local memory information, makes it difficult for models to tackle fine-grained EQA effectively.

Abrar \textit{et al.} \cite{anwar2024remembr} used retrieval-augmented LLM-agent to retrieve relevant memories by forming function calls and answer questions based on a real-time memory-building process.
Chen \textit{et al.} \cite{cheng2024efficienteqa} and Xie \textit{et al.} \cite{xie2024embodied} employed retrieval-augmented generation (RAG) to retrieve relevant images from accumulated observations and used a VLM for inference to generate responses.
However, planning that does not rely on memory information will lead to repetitive exploration, resulting in the agent's inability to answering questions correctly.
To this end, MemoryEQA establishes a comprehensive memory mechanism encompassing storage, update, and retrieval, enabling memory information to more effectively facilitate reasoning for efficient exploration and accurate answering.
\begin{figure*}[ht]
    \centering
    \includegraphics[width=\linewidth]{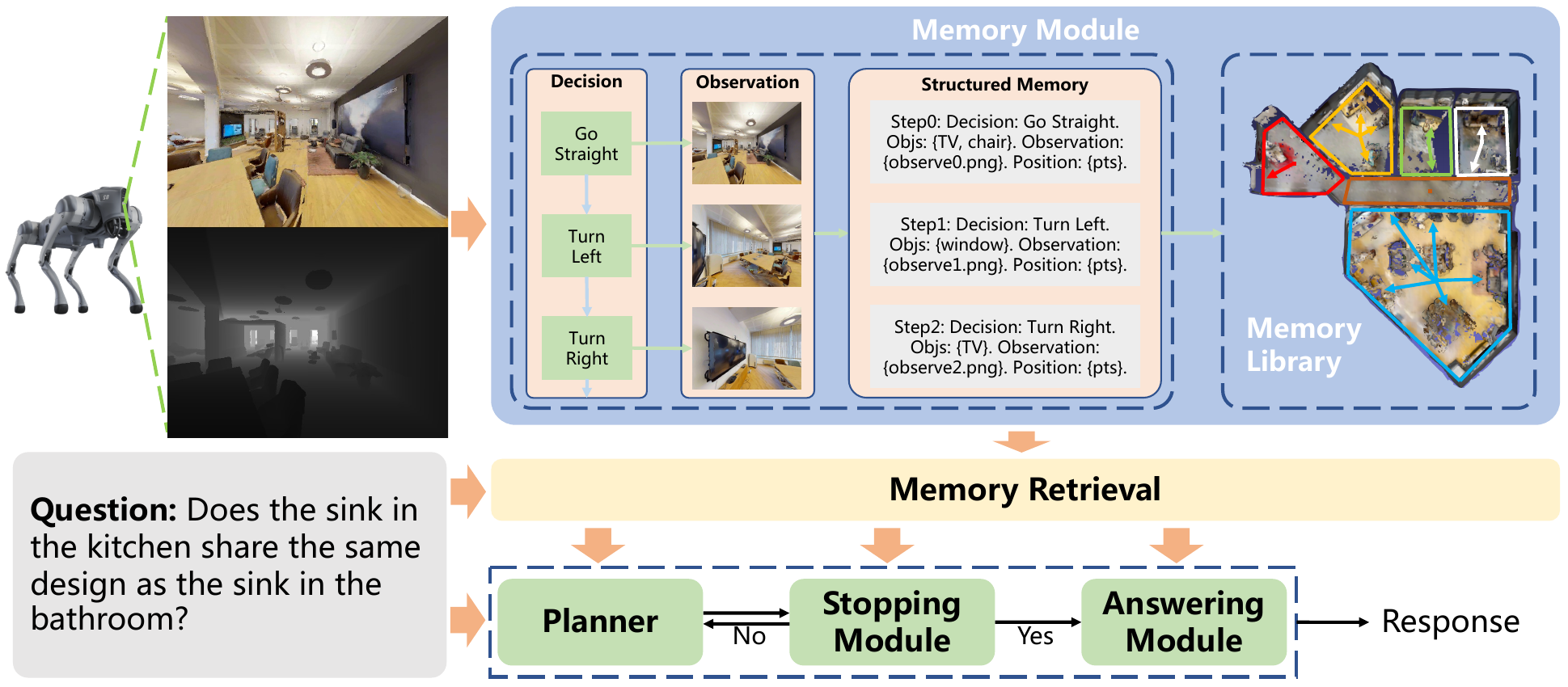}
    \caption{
    The overall framework of memory-centric EQA.
    Given the input question and agent observation in multiple steps, we gradually update the structural memory. 
    Subsequently, question-relevant memories are retrieved through the memory retriever and input into the respective modules.
    We concatenate the memories and the question, which are then fed into the planner to determine the next navigation direction . 
    When the agent arrives at a navigation point, it assesses whether to halt its journey. If the decision is `No', the process loops back to the memory update step; if `Yes', the agent responds to the question.
    }
    \label{fig:framework}
\end{figure*}

\section{Method}
MemoryEQA can efficiently handle complex tasks that involve multiple targets across different regions. Our goal is to provide proper memory information as inputs to various modules, rather than using memory only at the time of answering. 

\subsection{Background}
Typically, an EQA agent consists of four modules: a planner $\mathcal{F}_P$, a memory module $\mathcal{F}_M$, an answering module $\mathcal{F}_A$, and a stopping module $\mathcal{F}_S$. 
We define the observation sequence as $O=\{{o_i}\}_{i=0}^{N_t}$ and pose sequence as $P=\{p_i,r_i\}_{i=0}^{N_t}$ received by the agent, where $p_i$ and $r_i$ is the $i$-th position and rotation, $N_t$ represents the total number of exploration steps.
We use $\mathcal{M}=\{o_i,p_i,m_i\}_{i=0}^{N_t}$ represent the memory library.
The usual EQA framework can be formalized as
\begin{equation}
\left\{
\begin{aligned}
o_{i+1}&=\mathcal{F}_P(o_i),\quad if \ \ \mathcal{F}_S(o_i)=0, \\
r&=\mathcal{F}_A(q,\mathcal{M}),\quad else,
\\
\end{aligned}
\right.
\end{equation}
where $q$ is the question, and $r$ is the response of VLMs. $\mathcal{F}_S(\cdot)$ is an indicator function: if the current observation can answer the question, it outputs 1; otherwise, it outputs 0.
During the exploration process, the agent will determine whether it can answer the question based on the current observation. If $\mathcal{F}_S(o_i)=0$, it will continue to explore to obtain the next observation. Otherwise, it will directly answer the question.
However, this approach only utilizes historical information for the answering module $\mathcal{F}_A$, which may result in redundant or insufficient exploration, leading to suboptimal success rates.

\subsection{Memory-Centric EQA Framework}

We propose the memory-centric EQA (MemoryEQA) framework, as shown in Figure~\ref{fig:framework}, which formulates the memory module $\mathcal{F}_M = \{f_M^r, f_M^u\}$ consisting of a retrieval function ($f_M^r$) and an update function ($f_M^u$). This module converts the retrieved information into an appropriate form as input to each component.
We define the memory information as $\mathcal{M}=\{m_i\}_{i=0}^{N_s} =\{f_M^u(o_i)\}_{i=0}^{N_s}$, where  $o_i$ is the image observation, and $m_i$ the natural language memory.
MemoryEQA can be formulated as
\begin{equation}
\left\{
\begin{aligned}
o_{i+1}&=\mathcal{F}_P(o_i,f_M^r(q_p)),\quad if \ \ \mathcal{F}_S(o_i,f_M^r(q_s))=0, \\ 
r&=\mathcal{F}_A(o_i,f_M^r(q_a),q),\quad else,
\\
\end{aligned}
\right.
\end{equation}
where $q_p$, $q_s$, and $q_a$ denote the queries used to retrieve memory information required by the planner, stopping module, and answering module, respectively (details see the supplementary materials).

\subsubsection{Planner}


We extend the classical frontier-based exploration \cite{yamauchi1997frontier} with memory guidance, which integrates top-$p$ frontier selection and VLM-based reasoning.
The environment is represented as an occupancy grid map 
$\mathcal{G}=\{g_{i,j}\}$, where each cell is labeled as free, occupied, or unknown. A frontier cell is defined as a free cell adjacent to at least one unknown cell, and connected frontier cells are grouped into regions $\mathcal{D}=\{D_1,D_2,\cdots,D_N\}$.
Each frontier $D_i$ is evaluated by a utility function $U(D_i)=\lambda_1G(D_i)-\lambda_2C(D_i)$, where $G(\cdot)$
measures expected information gain and $C(\cdot)$ denotes navigation cost. $\lambda_1$ and $\lambda_2$  are weight coefficients used to balance the importance of information gain and navigation cost in target selection.
Instead of selecting only the frontier with the highest utility, we retain the top-$p$ candidates to preserve diverse exploration directions:
$$
\mathcal{D}^*={\text{Top-}p}_{D_i\in\mathcal{D}}(U(D_i)).
$$
Each selected frontier centroid is projected to pixel space using the camera model, from which a visual patch is extracted.
For frontier candidates, a query $q_p$ is used to retrieve memory information providing relevant past observations and semantics.
The VLM then evaluates the consistency between multiple directions of the current observation and retrieved memory information, producing relevance scores.
We ultimately chose the exploration direction with the highest relevance score.
This strategy enables the agent to balance efficient spatial coverage with semantic relevance, where frontier utility ensures broad exploration, and memory-augmented VLM reasoning prioritizes task-relevant goals.

\subsubsection{Memory Module}

The memory module has three mechanisms: storage, update, and retrieval. Storage and update are used to maintain the memory library, while retrieval is used to extract task-related memory information library for different modules. 
As shown in Figure~\ref{fig:framework}, when other modules (planner, stopping module, and answering module) are called, the first step is to retrieve required memory information and then make further inferences based on the information.

\noindent \textbf{Storage.}
The memory information is stored in dense vectors to facilitate efficient retrieval. 
We employ a unified encoder $e_c(\cdot)$ to extract multi-modal memory features $f_i=e_c(m_i)$. These feature vectors are then stored in a vector library $\mathcal{L}=\{L_i\}_{k=0}^{N_t}=\{<I_i,p_i,r_i,m_i,f_i>\}_{i=0}^{N_t}$ with each memory assigned a unique index $I_i$ for subsequent retrieval operations. 
In addition, once the agent completes a task, the memory accumulated in that scene is persistently stored in scene memory library $\mathcal{M}^s=\{\mathcal{M}_t\}^{N_s}_{t=0}$, where $N_s$ is the number of scenes. It serves as the initial memory repository for future tasks performed in the same environment.

\noindent \textbf{Update.}
During the exploration process, the agent may reveal new environmental changes that necessitate memory updates. So we design a memory update mechanism to effectively prevent erroneous triggering of updates. The specific rules are as follows:

\begin{itemize}[leftmargin=*]
    \item We check whether the agent enters a known region with
\begin{align}
    &\min_{m_j \in \mathcal{M}} \|p_i - p_j\|_2 < \beta_p, \\
    &\min_{m_j \in \mathcal{M}} \angle (r_i,r_j) < \beta_r,
\end{align}
where $\|\cdot\|_2$ is the Euclidean distance operator. $\angle(\cdot,\cdot)$ is the Euclidean angle operator. In other words, if the distance between the current position and all positions in memory is greater than the threshold $\beta_p$, and the angle between the current orientation and all orientations in memory is greater than the threshold $\beta_r$, we regard that the agent enters a new region.

\item The similarities between the current observation and all memory observations exceed a threshold. We calculate the similarity 
\begin{equation}
    SIM(o_i,o_j)=\alpha\cdot\text{SSIM}(o_i,o_j) + (1-\alpha)\cdot\text{sim}(f_i,f_j),
\end{equation}
by integrating both structural similarity and semantic similarity, where $(f_i,f_j)=(e_c(o_i),e_c(o_j))$ denotes the current observation feature $f_i$ and the history observation feature $f_j$, respectively. The $\text{SSIM}$ \cite{wang2004image} is structural similarity, and measures the perceptual similarity between two images by comparing their luminance, contrast, and structural patterns. The $\text{sim}(\cdot)$ is the cosine similarity.
    
\item The current observation's field of view is not restricted (black areas do not exceed half of the observation).
\end{itemize}

The memory update mechanism is triggered if and only if the three rules above are satisfied. Otherwise, the observation is regarded as entering a new region, and the memory is directly added to the memory library.
Once an update is triggered, we assign the index $I_i$ of the memory to be updated to the new memory, and then remove the original memory to complete the update process.

\noindent \textbf{Retrieval.}
We adopt a multi-modal retrieval approach. Given the initial observation $o_0$ at a scene and the user question $q$, we use the unified encoder $e_c(\cdot)$ to generate a combined feature 
\begin{equation}
    f_q = \text{Norm}(\operatorname{concat}(e_c(o_0), e_c(q))).
\end{equation}
Then, we use $f_q$ as the query to retrieve memory information related to the current scene from the scene memory library $\mathcal{M}^s$ via cosine similarity, forming the initial memory library $\mathcal{M}$.
The above process is performed only once, before the agent begins exploration.
At each exploration step, we retrieve from the memory library $\mathcal{M}$ for the memory information $\mathcal{M}^{cur}$ required by different modules. We formalize the retrieval process as 
\begin{equation}
    \begin{split}
        \mathcal{M}^{cur}_{\{p,s,a\}}=\text{top}_k(\{m_i | (\|e_c(q_{\{p,s,a\}}) - e_c(m_i)\|_2 \\< \alpha_e(1 + \text{Entropy}(e_c(q_{\{p,s,a\}})))) \\ 
        \cap \text{sim}(e_c(q_{\{p,s,a\}}),e_c(m_i)) > \alpha_s\}),
    \end{split}
\end{equation}
where $\alpha_e$ and $\alpha_s$ are the weight of similarity, and $\mathcal{M}^{cur}_{\{p,s,a\}}$ denote the memory information for the planner, stopping module, and answering module, respectively.
We use entropy to measure the uncertainty of the query features, making it an adaptively adjustable distance threshold. When the entropy is low (\emph{i.e.}, the query semantics are clear), we aim to match only very close memories to avoid noise; when the entropy is high (\emph{i.e.}, the query semantics are ambiguous), we allow a wider matching range, otherwise no relevant memories might be retrieved. Additionally, the cosine similarity between the memory and the query should exceed the threshold $\alpha_s$.


The size of $\mathcal{M}^{cur}_{\{p,s,a\}}$ significantly impacts model performance (see supplementary material). To address this, we employ a dynamic k-value mechanism that adaptively samples varying amounts of memory information for the $\text{top}_k$ selection according to the complexities of the user query $q$ and the initial observation $o_0$. Specifically, we implement an entropy-based linear adjustment strategy for the value of $k$, formulated as
\begin{equation}
k = \left\lceil k_{\min} + \beta \cdot \text{Entropy}(f_q) \right\rceil,
\end{equation}
where $k_{min}$ denotes the minimum allowable value of $k$, $\left\lceil \cdot \right\rceil$ is the operator of floor, $\beta$ is the weight of entropy. Intuitively, this means that complex queries and scenes will retrieve more memory information.

\begin{table*}[h]
    \centering
    \caption{The comparison between our proposed MT-HM3D and existed dataset. We assume that each question involves at least 1 region, even if the question does not explicitly mention any region.}
    \label{tab:dataset}
    \begin{tabular}{c|cccccc}
    \toprule
       Dataset  & Simulator & Source & Real Scenes & Objects & Region & Creation \\
    \midrule
       MT-EQA\cite{yu2019multi} & House3D & SUNCG & \xmark & 3.20 & 1.0 & Rule-Based \\
       EXPRESS-Bench\cite{jiang2025beyond} & Habitat & HM3D & \cmark & 3.21 & 1.09 & VLMs \\
       HM-EQA\cite{ren2024explore} & Habitat & HM3D & \cmark & 2.87 & 1.0 & VLMs \\
       Open-EQA\cite{majumdar2024openeqa} &Habitat & ScanNet/HM3D & \cmark & 2.53 & 1.0 & Manual \\
       \textbf{MT-HM3D (Ours)} & Habitat & HM3D & \cmark & \textbf{3.44} & \textbf{1.49} & VLMs \\
    \bottomrule
    \end{tabular}
\end{table*}


\begin{figure}[ht]
    \centering
    \includegraphics[width=\linewidth]{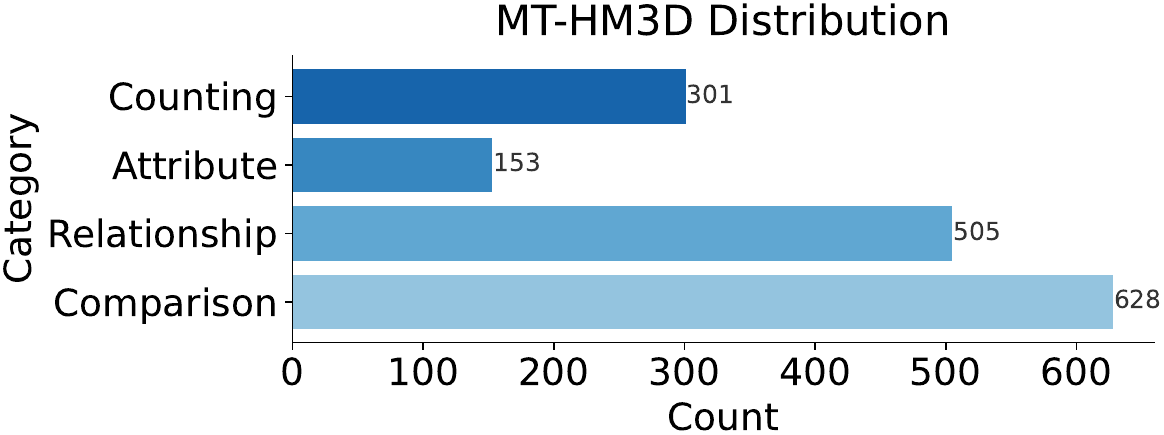}
    \caption{The category distribution of MT-HM3D.}
    \label{fig:distribution}
\end{figure}

\subsection{Stopping Criterion}
We combine the retrieved memory information with the current observation and prompt a VLM to obtain the current stopping confidence.
If the confidence exceeds a given threshold, the exploration stops, and the answering module is invoked. 
If the maximum number of steps is reached, the agent will also stop exploring, and the answering module is invoked. 
So the stopping module is formulated as 
$
\mathcal{F}_s=\mathbf{1}(\mathcal{P}(s|o_i,f_M^r(q_s)) > \gamma),
$
where $s$ is the confidence for stopping, $\mathbf{1}$ is an indicator function, which is 1 when the condition is met and 0 otherwise. $\gamma$ is the threshold.
In the implementation, we use a VLM as $\mathcal{P}$ and prompt it to provide the confidence in stopping the exploration process based on the retrieved memory information and the current observation. 
For detailed information, please refer to the supplementary material.

\subsection{Answering Module}
We employ a retrieval-augmented VLM as the answering module, which is invoked once the agent terminates exploration. Memory retrieval mechanism identifies and extracts only the memory content that is semantically relevant to the given question. This selective feeding strategy not only reduces computational overhead but also mitigates the risk of distraction from irrelevant or noisy information. Moreover, by grounding the VLM’s reasoning in the retrieved memory rather than raw observations, the answering process becomes more interpretable and contextually coherent, enabling the model to generate responses that faithfully reflect the agent’s past experiences and the spatial-temporal context of the environment.

\section{Benchmark Construction}
To evaluate the memory ability of EQA agents, we propose a multi-target EQA dataset based on HM3D \cite{ramakrishnan2021habitat}, namely MT-HM3D. Our MT-HM3D dataset includes 1587 QA samples on 500 scenes (more statistical data is in the supplemental material).

As shown in Table~\ref{tab:dataset}, the most significant distinction from existing datasets lies in the fact that the MT-HM3D constructs question-answering pairs targeting multiple regions and multiple objects in real-world scenarios, while exhibiting a richer diversity of question formats. 
We define four types of questions.
\begin{itemize}[leftmargin=*]
    \item Comparison: comparing the status of two or more entities, \textit{e.g.}, ``Is the bed closer to the chair than to the window in the bedroom?''
    \item Relationship: asking about the relationship between two entities, \textit{e.g.}, ``Are the red lampshade and the pipes in the same room?''
    \item Counting: requiring the agent to count an entity, \textit{e.g.}, ``How many chairs are around the dining table?''
    \item Attribute: assessing the difference in attributes between similar entities, \textit{e.g.}, ``Is the table in the kitchen bigger than the table in the study?''
\end{itemize}
The statistical details of the MT-HM3D dataset are illustrated in Figure~\ref{fig:distribution}.

We use GPT-4o \cite{gpt4o} and GPT-4-mini \cite{gpt4o} to construct multi-target question answering. The process consists of the following steps:
(1) Sample 5 to 10 images from the scene using the Habitat simulator.
(2) Input the sampled images and prompt GPT-4o \cite{gpt4o} to select multiple related objects from these images, specifying the output format to obtain unfiltered data.
(3) Use GPT-4-mini \cite{gpt4o} to filter the raw data, removing illogical question-answer pairs, and generate diverse QA data.
(4) Manually verify the correctness of the data to ensure that the questions can be answered within the embodied environment.
Specific prompts and more details are provided in the supplementary material.

\textbf{Metrics.} 
For multiple-choice questions, we make agents directly provide a predicted option, and then calculate the success rate (\%). 
For open-vocabulary questions, we use two metrics to evaluate the performance of agents: GPT score and ROUGE$_L$ \cite{lin2004rouge}. 
Then we use average normalization steps to evaluate the efficiency of the agents, and the specific calculation formula is
\begin{equation}
    Norm_{step}=\frac{1}{N}\sum_{i=0}^{N}\frac{N_i}{\sqrt{S_i} * \gamma_s},
\end{equation}
where $N$ means the number of samples, $N_i$ represents the explore step number of agents on the sample, $S_i$ is room size, and $\gamma_s$ is the relation ratio between max steps and room size. 
\section{Experiments}

\subsection{Implementation}
We use Qwen30VL-8B \cite{Qwen2.5-VL} and GPT-4o \cite{gpt4o} as the built-in VLMs, and employ Clip-ViT-Large \cite{radford2021learning} as the unified encoder to encode the multi-modal memory into 768-dimensional features. The object detector is YOLOv11\cite{khanam2024yolov11}. We use Faiss \cite{douze2024faiss} as a library of dense vectors.
Our simulator follows the setup of Explore-EQA \cite{ren2024explore} and conducts experiments in Habitat-Sim \cite{szot2021habitat}. Inference is performed on a single NVIDIA Tesla L40, with approximately 40 hours required to complete inference on the MT-HM3D dataset.

\begin{table*}
    \centering    
    \caption{Experiments on multiple methods across MT-HM3D, HM-EQA, and OpenEQA, testing various foundational models. $\dagger$ indicates the performance of the official implementation.}
    \small
    \begin{tabular}{c|c|c|cc|cc|ccc}
        \toprule
        \multirow{2}{*}{Exp.} & \multirow{2}{*}{Model} & \multirow{2}{*}{VLM} & \multicolumn{2}{c|}{MT-HM3D} & \multicolumn{2}{c|}{HM-EQA} & \multicolumn{3}{c}{OpenEQA} \\
        & & & Succ. $\uparrow$ & Step $\downarrow$ & Succ. $\uparrow$ & Step $\downarrow$ & GPT $\uparrow$ & ROUGE$_L$ $\uparrow$ & Step $\downarrow$ \\
        \midrule
        \multicolumn{10}{c}{Video LLM}  \\
        \midrule
        1 & \multirow{2}{*}{ExploreEQA\cite{ren2024explore}}  & VideoMind-7B\cite{liu2025videomind} & 35.13 & 0.52 & 51.3 & 0.61 & 31.43 & 0.28 & 0.61 \\
        2 &   & InternVideo2.5-8B\cite{wang2025internvideo2} & 39.58 & 0.53 & 53.52 & 0.59 & 33.14 & 0.3 & 0.59 \\
        \midrule
        3 & \multirow{2}{*}{MemoryEQA}  & VideoMind-7B\cite{liu2025videomind} & 41.93 & 0.46 & 54.44 & 0.58 & 31.34 & 0.29 & 0.58 \\
        4 &   & InternVideo2.5-8B\cite{wang2025internvideo2} & 41.15 & 0.46 & 53.36 & 0.55 & \textbf{33.69} & 0.31 & 0.58 \\
        \midrule
        \multicolumn{10}{c}{Spatial LLM}  \\
        \midrule
        5 & \multirow{2}{*}{ExploreEQA\cite{ren2024explore}}  & Spatialbot-3B\cite{cai2025spatialbot} & 31.53 & 0.53 & 49.61 & 0.71 & 25.97 & 0.15 & 0.61 \\
        6 &   & SpatialVLM-7B\cite{chen2024spatialvlm} & 33.78 & 0.55 & 53.98 & 0.61 & 30.5 & 0.25 & 0.7 \\
        \midrule
        7 & \multirow{2}{*}{MemoryEQA}  & Spatialbot-3B\cite{cai2025spatialbot} & 32.01 & 0.42 & 51.55 & 0.51 & 26.97 & 0.14 & 0.55 \\
        8 &   & SpatialVLM-7B\cite{chen2024spatialvlm} & 40.95 & 0.41 & 55.55 & 0.50 & 29.33 & 0.26 & 0.71 \\
        \midrule
        \multicolumn{10}{c}{Visual LLM} \\
        \midrule
        9 & \multirow{4}{*}{ExploreEQA\cite{ren2024explore}}  & Prismatic-7B$^\dagger$\cite{karamcheti2024prismatic} & 30.22 & 0.64 & 51.5 & 0.75 & 28.28 & 0.26 & 0.7 \\   
        10 & & Qwen2VL-7B\cite{wang2024qwen2}   & 33.18 & 0.76 & 44.31 & 0.74 & 27.36 & 0.24 & 0.74 \\
        11 & & Qwen3VL-7B\cite{Qwen2.5-VL}   & 33.43 & 0.7 & 50.13 & 0.69 & 31.29 & 0.26 & 0.71 \\
        12 & & GPT4o\cite{gpt4o}        & 33.21 & 0.64 & 47.4 & 0.77 & 29.21 & 0.26 & 0.77 \\
        \midrule
        13 & \multirow{3}{*}{MemoryEQA}   & Qwen2VL-7B\cite{wang2024qwen2}   & 40.4 & 0.4 & 55.9 & 0.52 & 27.64 & 0.28 & 0.6 \\
        14 & & Qwen3VL-7B\cite{Qwen2.5-VL}   & 41.95 & \textbf{0.4} & 56.65 & 0.49 & 30.53 & 0.3 & 0.61 \\
        15 &  & GPT4o\cite{gpt4o}        & \textbf{43.11} & 0.41 & \textbf{61.4} & \textbf{0.4} & 30.87 & \textbf{0.32} & \textbf{0.58} \\
        \bottomrule
    \end{tabular}
    \label{tab:main}
\end{table*}

\subsection{Main Results}
We compare MemoryEQA with ExploreEQA~\cite{ren2024explore} using three types of VLMs (Video LLMs~\cite{liu2025videomind,wang2025internvideo2}, Spatial LLMs~\cite{cai2025spatialbot,chen2024spatialvlm}, and Multi-Image LLMs~\cite{wang2024qwen2,Qwen2.5-VL,gpt4o,karamcheti2024prismatic}) and evaluate them on MT-HM3D, HM-EQA~\cite{ren2024explore}, and OpenEQA~\cite{majumdar2024openeqa}.

On MT-HM3D, MemoryEQA attains a success rate of 43.11\%, outperforming baseline by 9.9\% (Exp.12 vs Exp.15), highlighting the critical role of hierarchical memory in multi-target tasks. 
By comparing Exp.1-2 and Exp.9-12, it can be observed that the VideoLLM-based EQA method shows an improvement on MT-HM3D (ranging from 1.92\% to 9.36\%), as the video contains historical information. However, due to the excessive presence of irrelevant information within the video memory, which interferes with the final answer, the performance is suboptimal. Comparing Exp.1-2 and Exp.3-4, it is evident that using the MemoryEQA framework leads to further improvements, indicating that the memory structure in MemoryEQA is more suitable for the EQA task.
Furthermore, by comparing Exp.5-6 and Exp.9-12, it can be concluded that the Spatial LLM provides limited benefits for the EQA task. This may be due to the scarcity of questions related to spatial relationships (e.g., "What is the size of the sofa?") in the dataset, which prevents the Spatial LLM from fully exploiting its potential, resulting in its performance degrading to that of a standard LLM.
It sum up, MemoryEQA exhibits superior performance in multi-modal reasoning tasks, particularly in complex scene understanding and knowledge integration.

The analysis of experimental results across Exp.3-4, Exp.7-8, and Exp.13-15 reveals a significant positive correlation between the performance of the VLM and the effectiveness of the EQA system. This observation underscores the critical role that VLM plays in enhancing the EQA system's ability to process and interpret complex queries within visual environments. As the VLM's accuracy and understanding improve, so does the EQA system's capacity to deliver precise and contextually relevant answers, demonstrating a synergistic relationship between the two components. This finding highlights the importance of advancing VLM capabilities to further boost the overall performance of EQA systems in practical applications.

\begin{table}[ht]
    \centering
    \caption{Experiment on the ablation of memory strategies. (U: Memory update strategy. K: K-value selection strategy. R: Memory retrieval strategy.)}
    \begin{tabular}{l|cc|cc}
    \toprule
    \multirow{2}{*}{Strategy} & \multicolumn{2}{c|}{MT-HM3D} & \multicolumn{2}{c}{HM-EQA} \\
     & Succ. $\uparrow$ & Step $\downarrow$ & Succ. $\uparrow$ & Step $\downarrow$ \\
    \midrule
    None      & 33.18 & 0.66 & 44.31 & 0.74 \\
    U         & 33.41 & 0.67 & 44.29 & 0.71 \\
    U + R     & 39.69 & 0.45 & 53.16 & 0.53 \\
    U + R + K & 41.95 & 0.4 & 56.65 & 0.49 \\
    \bottomrule
    \end{tabular}
    \label{tab:strategies}
\end{table}

\begin{table}[ht]
    \centering
    \caption{Ablation study on MT-HM3D and HM-EQA to evaluate the impact of the memory information into different modules on performance. (S: Stop Criterion, A: Answering Module, P: Planner)}
    \begin{tabular}{l|cc|cc}
    \toprule
    \multirow{2}{*}{Module} & \multicolumn{2}{c|}{MT-HM3D} & \multicolumn{2}{c}{HM-EQA} \\
     & Succ. $\uparrow$ & Step $\downarrow$ & Succ. $\uparrow$ & Step $\downarrow$ \\
    \midrule
    None      & 30.22 & 0.64 & 51.55 & 0.75 \\
    S         & 35.1 & 0.64 & 52.24 & 0.7 \\
    S + A     & 40.99 & 0.61 & 54.76 & 0.71 \\
    S + A + P & \textbf{41.95} & \textbf{0.4} & \textbf{56.65} & \textbf{0.49} \\
    \bottomrule
    \end{tabular}
    \label{tab:module}
\end{table}


\begin{figure*}[h]
    \centering
    \includegraphics[width=1.0\linewidth]{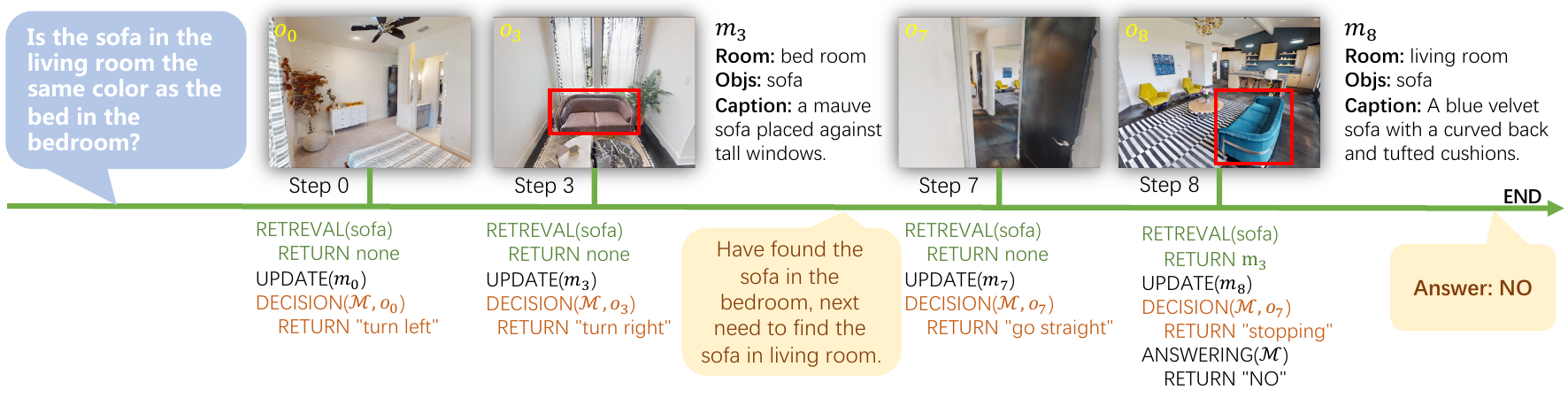}
    \caption{Visualization of the exploration process. Memory plays a pivotal role in embodied question answering, enabling context-aware decision-making across various environments. 
    $\texttt{RETREVAL}(\cdot)$ represents retrieving relevant memories from the memory library $\mathcal{M}$.
    $\texttt{UPDATE}(\cdot)$ represents updating the memory obtained from the current observation to the memory library $\mathcal{M}$.
    $\texttt{DECISION}(\cdot,\cdot)$ represents determining the next exploration direction based on memory and current observation.
    $\texttt{ANSWERING}(\cdot)$ represents answering based on memory.
    $m_i$ of the observation image represents the memory information under current observation $o_i$ (simplified version)
    }
    \vspace{-0.5em}
    \label{fig:qualitative}
\end{figure*}

\subsection{Ablation Studies}
As shown in Table~\ref{tab:strategies}, we evaluated different strategies' impact on MemoryEQA performance. The k-value selection depends on memory retrieval: without retrieval, all memories are used as input, and without k-value selection, k is fixed at 5. Initially, the baseline model without memory strategies performs poorly, indicating suboptimal memory management. Introducing memory update (U) shows slight improvement, but its effect on success and efficiency is limited, likely due to low memory update needs in static scenarios.
When memory retrieval (R) is combined with memory update (U+R), significant improvements are observed, enabling the system to access more relevant past experiences, thus boosting performance. The most significant gains occur when all three strategies are applied, with k-value selection enhancing retrieval by removing redundant information and selecting relevant data.
Overall, combining memory update, retrieval, and k-value selection leads to the most efficient memory management.

As shown in Table~\ref{tab:module}, the ablation study evaluates the impact of the memory information into various module on MT-HM3D and HM-EQA datasets. Without memory (None), baseline success rates are 30.22\% (MT-HM3D) and 51.5\% (HM-EQA), with suboptimal step efficiency. Introducing the Stop Criterion (S) alone yields marginal improvements in MT-HM3D (+0.69\% Succ., reduced steps).
Combining Stopping and Answering Module (S+A) significantly boosts performance: MT-HM3D’s success rises by 10.77\% (to 40.99\%), and HM-EQA improves by 3.21\% (to 54.76\%), confirming that memory-enhanced answering critically resolves ambiguities in both tasks.
Full integration (S+A+P) achieves peak results: MT-HM3D reaches 41.95\% Succ. (11.73\% gain over baseline) with a 0.4 norm step reduction, while HM-EQA attains 56.65\% success rate (+5.1\%) and fewer steps (0.49). 
The planner's contribution is crucial, in synergy with S and A to optimize task decomposition and trajectory planning. 
This demonstrates that memory benefits all modules cumulatively, with the planner driving efficiency by mitigating redundant exploration.


\subsection{Visualization}
Figure~\ref{fig:qualitative} illustrates the critical role of memory in embodied question answering (EQA) systems. In EQA tasks, an agent interacts with its environment while simultaneously processing information to answer questions. Memory is essential for retaining key details, such as the location of objects or the layout of the environment, which directly informs the agent’s decisions and actions. In the example shown, the agent uses its memory to track the positioning of objects—such as the sofa in different rooms—allowing it to answer the question of whether the sofas in the living room and bedroom are the same color. By leveraging memory, the system can make informed, context-sensitive decisions, navigating the environment and adjusting its actions to retrieve and analyze relevant information. This capability is fundamental for improving the accuracy and efficiency of embodied question answering tasks in dynamic, real-world scenarios.

\section{Conclusion}
In this paper, we propose a memory-centric EQA framework, named MemoryEQA. MemoryEQA establishes a comprehensive memory management strategy that includes storage, updating and retrieval, enabling memory information to more efficiently and accurately assist other modules in reasoning.
Additionally, we have constructed a benchmark MT-HM3D based on the real-world HM3D scenes. 
This benchmark includes four categories of complex tasks, allowing for comprehensive validation of the memory capabilities of EQA models. 
The effectiveness of MemoryEQA has been demonstrated through performance on the MT-HM3D, HM-EQA and Open-EQA datasets. 
{
    \small
    \bibliographystyle{ieeenat_fullname}
    \bibliography{main}
}

\clearpage
\setcounter{page}{1}
\maketitlesupplementary

\section{Method Details}
$q_p$, $q_s$, and $q_a$ denote the queries used to retrieve memory information required by the planner, stopping module, and answering module, respectively. The specific form of the query is as follows
\begin{lstlisting}[language=Python, caption={The query $q_p$ used by planner.}, label=lst:scene_cap]
Retrieve memory entries describing the spatial layout and previously observed regions near the current viewpoint.
Current viewpoint: <agent_pose>
Visible objects: <object_list>
Goal type: <goal_type>
Return memory that contains navigable regions, unexplored frontiers, and previously seen locations relevant to planning the next movement.
\end{lstlisting}
\begin{lstlisting}[language=Python, caption={The query $q_s$ used by stopping module.}, label=lst:scene_cap]
Retrieve memory entries related to the existence, location, or appearance of the target object or target region.
Current observation: <observation_text>
Target specification: <question_text>
Return memory that may indicate the target has been found or is likely within the currently observed region.
\end{lstlisting}
\begin{lstlisting}[language=Python, caption={The query $q_a$ used by answering module.}, label=lst:scene_cap]
Retrieve memory entries that contain semantic information relevant to answering the question.
Question: <question_text>
Current observation: <observation_text>
Return memory that includes object attributes, spatial relations, and scene facts related to the queried concept.
\end{lstlisting}

\begin{figure}
  \centering
  \includegraphics[width=0.38\textwidth]{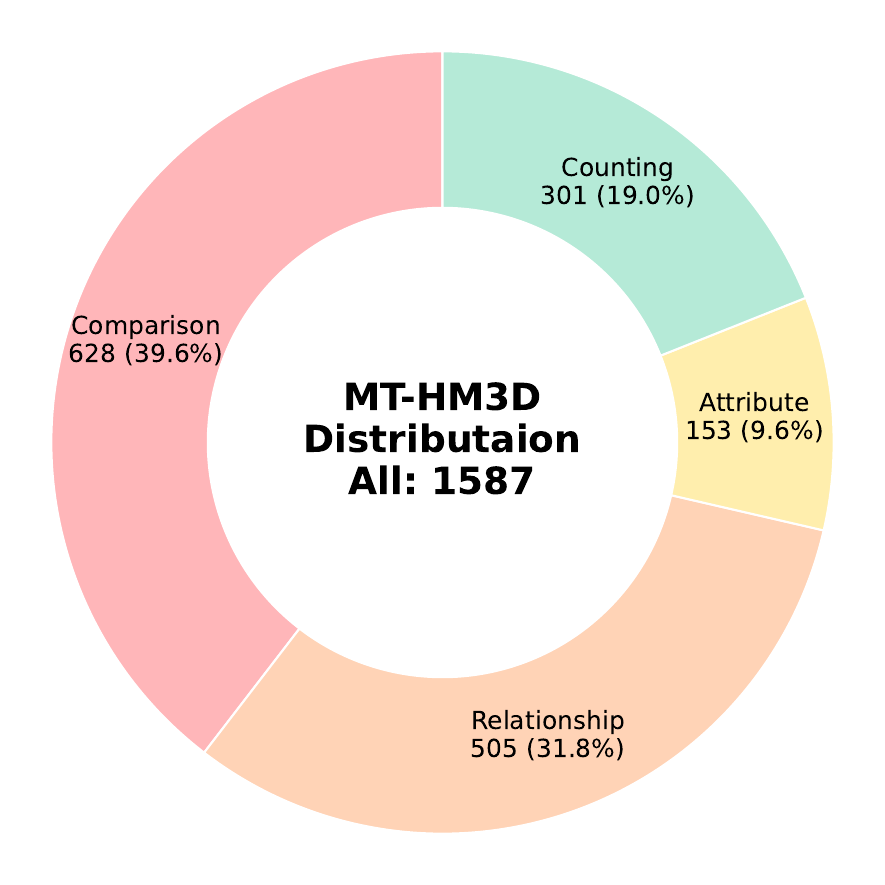}  
  \caption{Statistic of our proposed MT-HM3D.}
  \label{fig:distribution}
\end{figure}
We utilize prompt~\ref{lst:scene_cap} and prompt~\ref{lst:obj_cap} to prompt VLM to generate coarse-grained scene description and fine-grained object description respectively.
\begin{lstlisting}[language=Python, caption={Prompt for coarse-grained scene description.}, label=lst:scene_cap]
Describe this image. Output in the following format:
```
Room: <room>
Object: <obj1>, <obj2>,...
Description: <description>
```
\end{lstlisting}

\begin{table*}[ht]
    \centering
    \caption{The comparison between our proposed MT-HM3D and existed dataset.}
    \label{tab:dataset}
    \begin{tabular}{c|cccccc}
    \toprule
       Dataset  & Simulator & Source & Real Scenes & Objects & Region & Creation \\
    \midrule
       EQA-v1\cite{das2018embodied} & House3D & SUNCG & \xmark & - & - & Rule-Based \\
       MP3D-EQA\cite{wijmans2019embodied} & MINOS & MP3D & \cmark & - & - & Rule-Based \\
       MT-EQA\cite{yu2019multi} & House3D & SUNCG & \xmark & 3.20 & 1.0 & Rule-Based \\
       IQA\cite{gordon2018iqa} & AI2THOR & - & \xmark & - & - & Rule-Based \\
       EXPRESS-Bench\cite{jiang2025beyond} & Habitat & HM3D & \cmark & 3.21 & 1.09 & VLMs \\
       HM-EQA\cite{ren2024explore} & Habitat & HM3D & \cmark & 2.87 & 1.0 & VLMs \\
       Open-EQA\cite{majumdar2024openeqa} &Habitat & ScanNet/HM3D & \cmark & 2.53 & 1.0 & Manual \\
       \textbf{MT-HM3D (Ours)} & Habitat & HM3D & \cmark & \textbf{3.44} & \textbf{1.49} & VLMs \\
    \bottomrule
    \end{tabular}
\end{table*}

\begin{lstlisting}[language=Python, caption={Prompt for fine-grained object description.}, label=lst:obj_cap]
Please describe the category, arrtibute, and description of the object.
Output in the following format:
```
cate: [category]
attr: [arrtibute]
desc: [description]
```
\end{lstlisting}
For the stopping module, we prompt the VLM to evaluate the confidence of the response through a multiple-choice question format. 
If the model selects a confidence level of C or higher, it will cease exploration and proceed to provide the answer.
\begin{lstlisting}[language=Python, caption={Prompt for the evaluation of confidence.}]
Consider the question: `{Question}'. How confident are you in answering this question from your current perspective?
A. Very low
B. Low
C. Medium
D. High
E. Very high
Answer with the option's letter from the given choices directly.
\end{lstlisting}

Prompt~\ref{lst:close_response} and prompt~\ref{lst:open_response} are used to prompt the answering module to closed-vocabulary response and open-vocabulary response.
\begin{lstlisting}[language=Python, caption={Prompt for the response.}, label=lst:close_response]
{Question} Answer with the option's single letter from the given choices directly.
\end{lstlisting}

\begin{lstlisting}[language=Python, caption={Prompt for the response on open-vocabulary data.}, label=lst:open_response]
{Question} Answer with the brief sentence.
\end{lstlisting}

Next, prompt~\ref{lst:local_sem} and prompt~\ref{lst:global_sem} follow ExploreEQA~\cite{ren2024explore} setting to prompt VLM to output explore direction.
\begin{lstlisting}[language=Python, caption={Prompt for the explore direction.}, label=lst:local_sem]
Consider the question: '{Question}', and you will explore the environment for answering it.
Which direction (black letters on the image) would you explore then? Provide reasons and answer with a single letter.
\end{lstlisting}

\begin{lstlisting}[language=Python, caption={Prompt for the Exploring Value.}, label=lst:global_sem]
Consider the question: '{Question}', and you will explore the environment for answering it. 
Is there any direction shown in the image worth exploring? Answer with Yes or No.
\end{lstlisting}

\section{Data Construction}

\subsection{Statistics}
To evaluate the memory capabilities of EQA models, we propose an EQA dataset that features multiple targets distributed across various regions. As shown in Table ~\ref{tab:dataset}, our dataset exhibits the highest average number of targets and regions, significantly increasing the complexity of the task. The requirement to explore multiple targets across different regions poses a substantial challenge to the model's memory capacity. The statistical details of the MT-HM3D dataset are illustrated in Figure ~\ref{fig:distribution}.

\subsection{Metrics}
For multiple-choice questions, we have the model directly provide a predicted option, and then calculate the success rate (\%). 
For open-vocabulary questions, we use two metrics to evaluate performance of the model: GPT score and ROUGE$_L$ \cite{lin2004rouge}. 
The specific calculation methods are as follows:
\begin{align}
    &ROUGE_L=\frac{2LCS(C,R)}{|R|+|C|}, \\
    &S_{GPT}=GPT4o(promtp,C,R),
\end{align}

where $LCS(\cdot,\cdot)$ is longest common subsequence, $C$ is prediction sentence, $R$ is ground-truth sentence, $cos$ is cosine similarity, $|\cdot|$ is calculating the number of words.

And then we use average normalization steps to evaluate the efficiency of the agents.
The specific calculation formula is
\begin{equation}
    Norm_{step}=\frac{1}{N}\sum_{i=0}^{N}\frac{N_i}{\sqrt{S_i} * \gamma_s},
\end{equation}
where $N$ means the number of samples, $N_i$ represent the explore step number of agent on current sample, $S_i$ is room size, and $\gamma_s$ is the relation ratio between max steps and room size. 

\begin{table*}
    \centering
    \caption{All hyper-parameters list.}
    \label{tab:hyperparameter}
    \scalebox{0.95}{
    \begin{tabular}{llll}
    \toprule
    \textbf{Hyperparameter} & \textbf{Value} & \textbf{Hyperparameter} & \textbf{Value} \\
    \midrule
    
    seed & 42 &
    device & cuda \\
    \midrule
    \textbf{vlm} \\
    model name or path & Qwen2-VL-7B \\
    
    \midrule
    \textbf{rag} \\
    use rag & true &
    text &  clip-vit-large-patch14 \\
    visual & clip-vit-large-patch14 &
    dim & 1536 \\
    max retrieval num & 10 &
    $\beta$ & 1.0 \\
    $\beta_p$ & 5.0 &
    $\beta_r$ & 15.0 \\
    $\alpha$ & 0.5 &
    $\alpha_e$ & 1.0 \\
    
    \midrule
    \textbf{camera/image} \\
    detector & yolo11x.pt &
    camera height & 1.5 \\
    camera tilt deg & -30 &
    img width & 640\\
    img height & 480 &
    hfov & 120 \\
    tsdf grid size & 0.1 &
    margin w ratio & 0.25 \\
    margin h ratio & 0.6 \\
    
    \midrule
    \textbf{navigation} \\
    init clearance & 0.5 &
    max step room size ratio & 3 \\
    black pixel ratio & 0.7 &
    min random init steps & 2 \\
    
    \midrule
    \textbf{planner} \\
    dist T & 10 &
    unexplored T & 0.2 \\
    unoccupied T & 2.0 &
    val T & 0.5 \\
    val dir T & 0.5 &
    max val check & 3 \\
    smooth sigma & 5 &
    eps & 1 \\
    min dist from cur & 0.5 &
    max dist from cur & 3 \\
    frontier spacing & 1.5 &
    min neighbors & 3 \\
    max neighbors & 4 &
    max unexplored & 3 \\
    max unoccupied & 1 \\
    
    \midrule
    \textbf{visual prompt} \\
    cluster threshold & 1.0 &
    num prompt points & 3 \\
    num max unoccupied & 300 &
    min points clustering & 3 \\
    point min dist & 2 &
    point max dist & 10 \\
    cam offset & 0.6 &
    min prompt points & 2 \\
    circle radius & 18 \\
    \bottomrule
    
    \end{tabular}}
\end{table*}

\begin{figure*}[h]
    \centering
    \includegraphics[width=\linewidth]{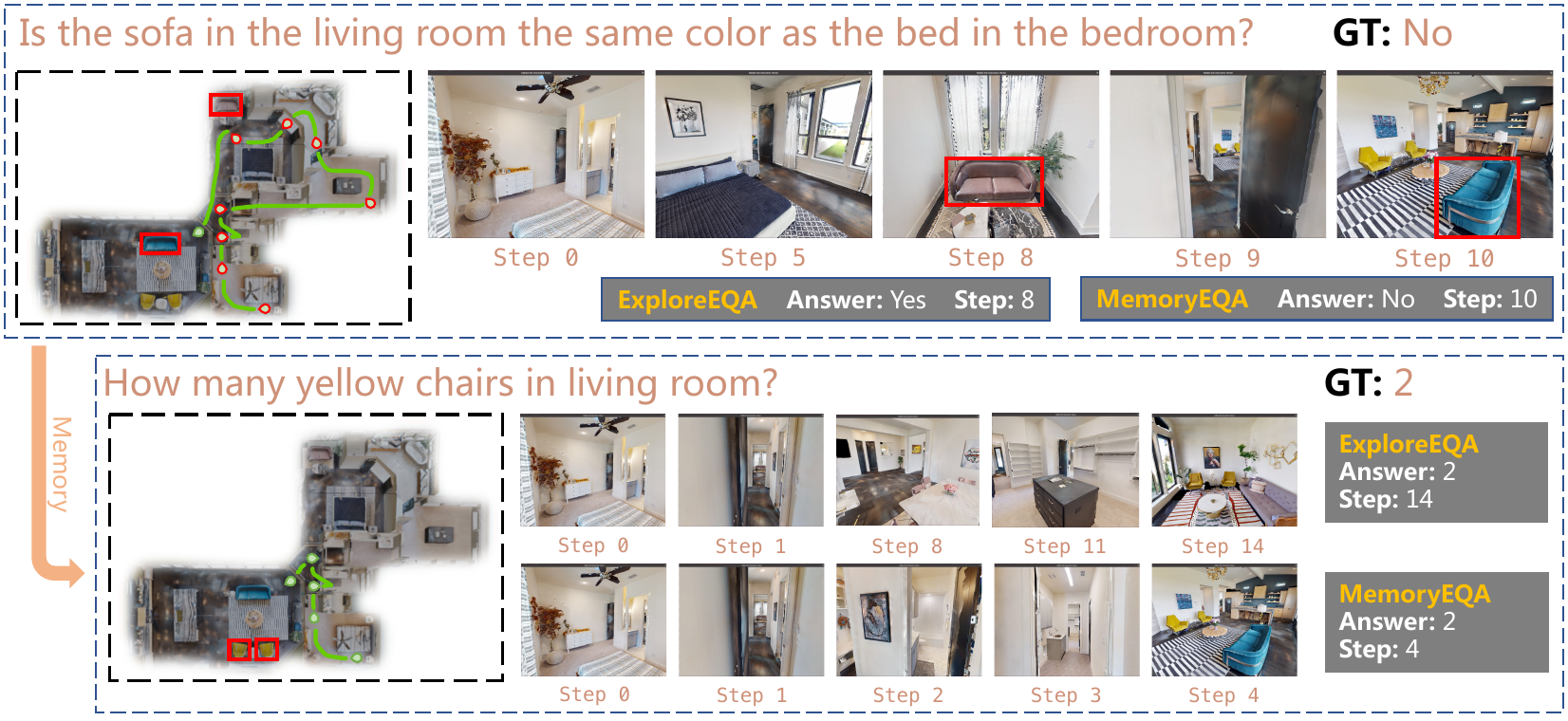}
    \caption{Visualization of Two-Round QA in the HM3D Dataset. The figure above illustrates the performance of the first round of QA baseline and MemoryEQA, with red dots representing the planning points of ExplorEQA and green lines indicating the planning path of MemoryEQA. The figure below demonstrates the results of the second round of QA, where the information collected during the exploration process of the first round is used as memory for further planning.}
    \label{fig:visualization}
    \vspace{-1em}
\end{figure*}

\subsection{Prompts}
We sample multiple images from the scene, stitch them together, and input the images along with a prompt~\ref{lst:data_gen} to generate proposed question-answer pairs.
\begin{lstlisting}[language=Python, caption={Prompt for data generation.}, label=lst:data_gen]
Select one or some object from those image (they must be related objects). 
Now, I hope you POSE A QUESTION based on the relationship between these two objects, and the question must meet the following requirements:
1. Select the question label from the following list: [Comparison, Relationship, Counting, Attribute];
1. do not ask "why" and "how"; 
2. the names of the objects must appear in the question;
3. avoid descriptions such as "left image" or "right image";
4. provide four candidate answer options as well as the correct answer option.

For example:
Does <OBJ1> share same color as <OBJ2> in <ROOM>?
Does <OBJ1> in <ROOM1> share same color as <OBJ2> in <ROOM2>?
Is <OBJ1> bigger/smaller than <OBJ2> in <ROOM>?
Is <OBJ1> in <ROOM1> bigger/smaller than <OBJ2> in <ROOM2>?
Is <OBJ1> closer than/farther from <OBJ2> than <OBJ3> in <ROOM>?
Is <ROOM1> bigger/smaller than <ROOM2> in the house?
How many <OBJ1> in <ROOM1>?
How many rooms have <OBJ1>?

The output must strictly following the format below:
```
Objects: [object1]; [object2]
Question: [question]
Options: [A. xxx; B. xxx; C. xxx; D. xxx]
Answer: [answer]
Label: [label]
```
\end{lstlisting}
After obtaining the question-answer pair, we use prompts~\ref{lst:data_verify} to verify the validity of the question and answer.
\begin{lstlisting}[language=Python, caption={Prompt for Verify the rationality of the data.}, label=lst:data_verify]
I need you to help me check whether the question meets my requirements, which are as follows:
(1) The question requires additional visual input to answer;
(2) The objects mentioned in the question can be found in an indoor scene;
(3) For an exploration robot in any location, the references in the question are clear during the robot's autonomous exploration process.
The question is as follows: {}
Please help me check whether the question meets the requirements. If it does not meet the requirements, answer "No" and provide a reason; if it meets the requirements, simply answer "Yes."
\end{lstlisting}

\begin{table}
    \centering
    \caption{The impact of memory retrieval quantity on model performance.}
    \begin{tabular}{c|cc}
        \toprule
           K  & Succ. $\uparrow$ & Step $\downarrow$ \\
        \midrule
           0   & 34.3 & 0.7 \\
           2   & 44.3 & 0.45 \\
           4   & \textbf{49.65} & \textbf{0.4} \\
           8   & 48.7 & 0.42 \\
           12  & 46.21 & 0.3 \\
        \bottomrule
        \end{tabular}
	\label{tab:k}
\end{table}

\begin{figure}[h]
    \centering
    \includegraphics[width=\linewidth]{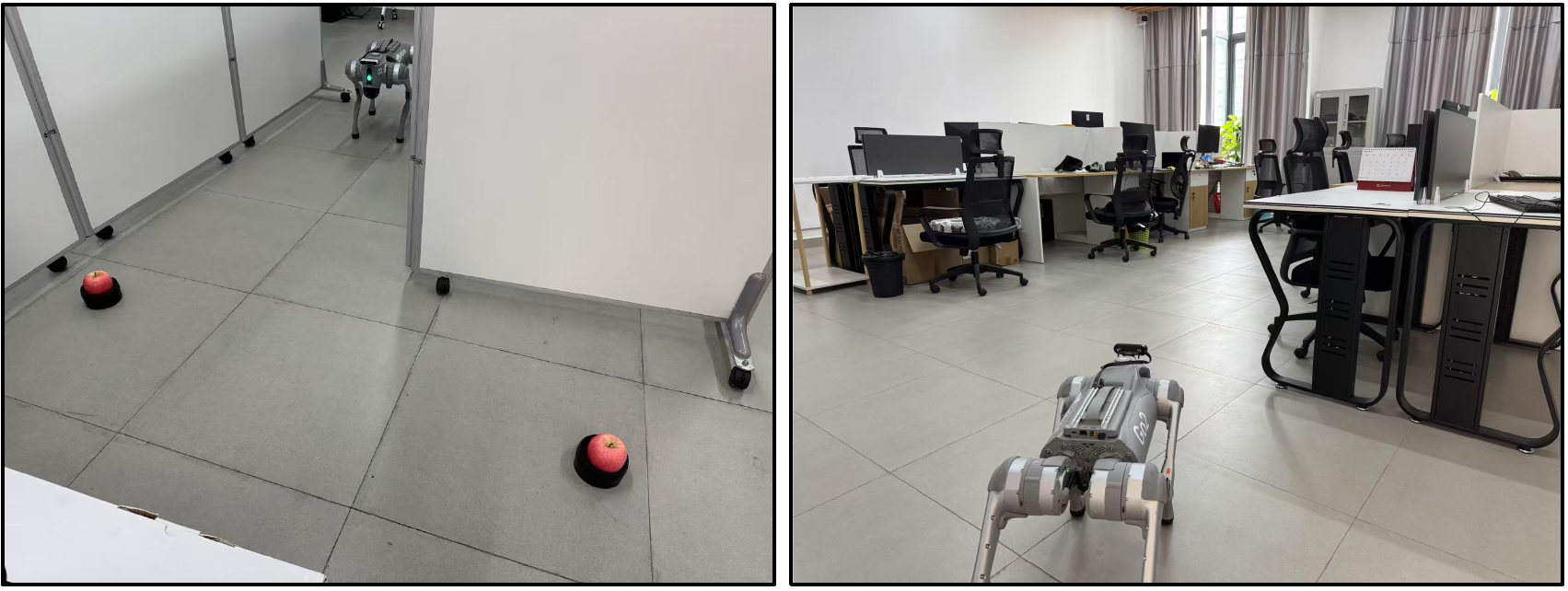}
    \caption{Experiment on real world. Left: handmade scenario. Right: office}
    \label{fig:realworld}
\end{figure}

\section{More Experiments}

\subsection{Implementation Details}
We have listed all the hyper-parameters used for our best results in Table ~\ref{tab:hyperparameter}.

\subsection{Parameter Analysis}
As shown in Table ~\ref{tab:k}, the selection of the K value is a pivotal factor, as both excessively large and unduly small K values can precipitate suboptimal performance. 
An insufficiently small K value engenders a dearth of memorized information, whereas an inordinately large K value not only induces informational redundancy but also fosters hallucinations in the response module.
We believe that using different K values for different problems may be a more optimal strategy. Specifically, a dynamic memory window, rather than a fixed K value, could better adapt to varying complexities, which warrants further investigation.

\subsection{Real-World Results}
We conducted experiments using the Unitree Go2 robot in both office settings and artificially constructed scenarios. The RGB-D images were captured using the RealSense D435i depth camera, with DDS serving as the underlying middleware, and the control of the agent was accomplished on the ROS2 platform \cite{macenski2022robot}. The multi-modal large model was deployed on 8 * Nvidia Tesla L40.

As shown in Figure ~\ref{fig:realworld}, 
in the handmade scenario, we posed an open-vocabulary question to the agent: ``How many apples are there in the scene?" After a thorough exploration, the agent provided an answer and did not recount when revisiting previously explored areas.
In the office setting, we inquired, ``Is the black garment placed on the chair?" The agent swiftly deduced the correct answer following a simple planning process.

\subsection{Visualization}
As shown in Figure ~\ref{fig:visualization}, we have visualized ExploreEQA and MemoryEQA to demonstrate the advantages of our global and local memory mechanisms. The figure above illustrates an example of local memory. The baseline method provided an incorrect answer immediately after observing a sofa, whereas MemoryEQA, upon observing the first sofa, did not retrieve information about another sofa from its memory and thus continued its exploration, ultimately delivering the correct response.

We stored information collected during the exploration process depicted in the figure above as persistent memory and input it into the question-answering process shown in the figure below. Since the yellow sofa was observed at the 10-th step during the first question, the agent, upon revisiting the scene, could prioritize exploration based on the coordinates of the yellow sofa observed previously, requiring only four steps to provide the correct response. In contrast, the baseline method took 14 steps of exploration to arrive at the correct answer.

\end{document}